\documentclass{article}
\usepackage{amssymb}
\usepackage{graphicx}
\usepackage{amsmath}
\usepackage[preprint]{corl_2025} 

\newcommand{\email}[1]{\href{mailto:#1}{\nolinkurl{#1}}}

\newcommand{\fig}[1]{Fig.~\ref{#1}}

\newcommand{\titlelong}[0]{SemanticFeels: Semantic Labeling\\ during In-Hand Manipulation}

\title{\titlelong}

\author{
  Anas Al Shikh Khalil \\
  LASR Lab\\
  TU Dresden
   \And
  Haozhi Qi \\
  UC Berkely \\
 \AND
  Roberto Calandra \\
  LASR Lab\\
  TU Dresden \\
}

\begin{document}
\maketitle


\begin{abstract}
As robots become increasingly integrated into everyday tasks, their ability to perceive both the shape and properties of objects during in-hand manipulation becomes critical for adaptive and intelligent behavior. We present SemanticFeels, an extension of the NeuralFeels framework that integrates semantic labeling with neural implicit shape representation, from vision and touch. To illustrate its application, we focus on material classification: high‑resolution Digit tactile readings are processed by a fine‑tuned EfficientNet‑B0 convolutional neural network (CNN) to generate local material predictions, which are then embedded into an augmented signed distance field (SDF) network that jointly predicts geometry and continuous material regions. Experimental results show that the system achieves a high correspondence between predicted and actual materials on both single- and multi-material objects, with an average matching accuracy of 79.87\% across multiple manipulation trials on a multi-material object.
\end{abstract}

\keywords{Tactile Sensing, Dexterous Manipulation, Semantic Labeling} 


\section{Introduction}
\label{sec:introduction}
As robots become more prevalent in our daily lives, their ability to perceive and understand the physical properties of objects with which they interact becomes ever more critical. The success of a robot-environment interaction often depends on the robot’s ability to navigate its surroundings, identify objects, and perform tasks efficiently and safely. To achieve this, robots must not only track the pose and map the shape of objects — a process commonly referred to as Simultaneous Localization and Mapping (SLAM) — but must also associate high-level qualitative features known as semantic features, such as physical properties, material composition, and surface roughness. This semantic understanding provides a valuable context that can help robots adapt their manipulation strategies based on the specific properties of the objects they encounter.

\begin{figure}
    \centering
    \includegraphics[width=1\linewidth]{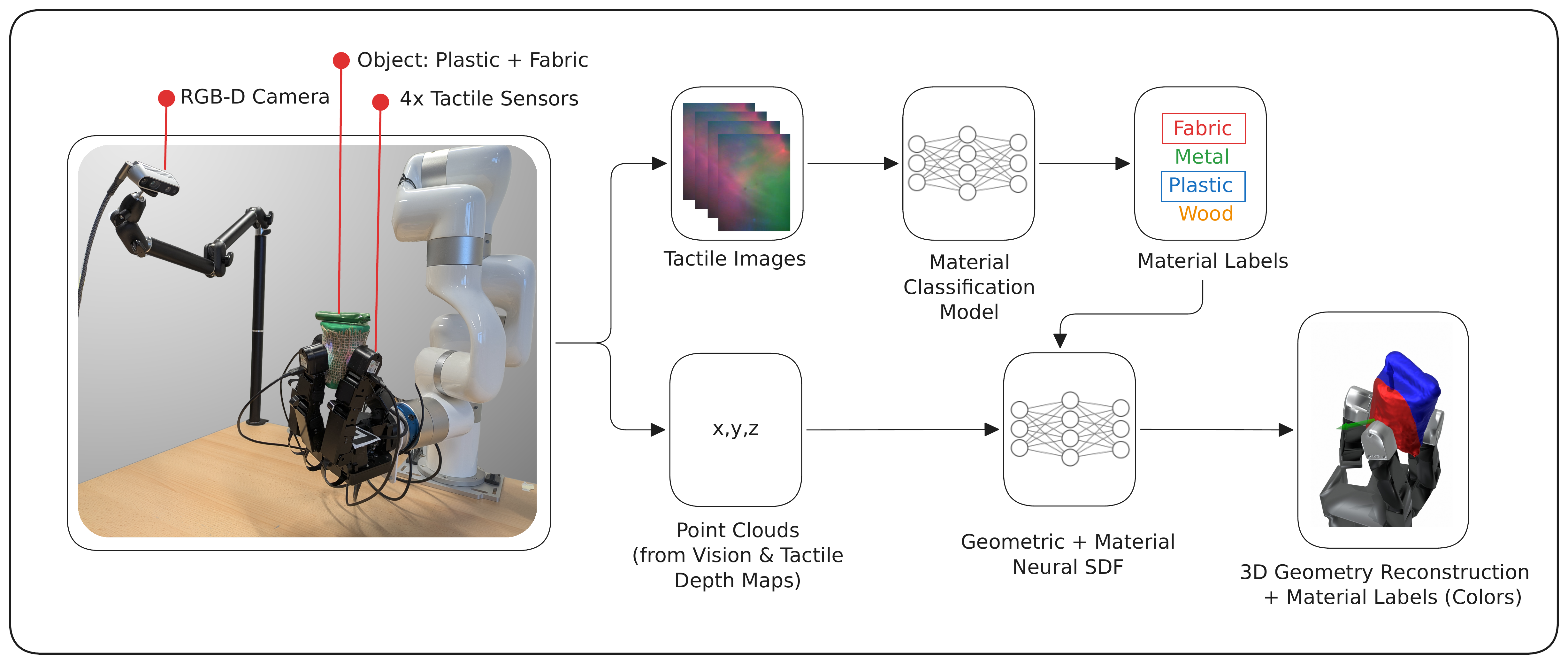}
    \caption{SemanticFeels framework: Diagram of our hardware and backend pipeline. An RGB-D camera and multiple tactile sensors collect visual and touch data from objects. This information is used to generate point clouds. Tactile images obtained from the tactile sensors, are processed by a classification model to predict local material types. These predictions are fused into a neural signed distance field (SDF) that reconstructs the object’s 3D shape while assigning semantic material labels.}
    \label{fig:overview}
\end{figure}

Robotic in-hand manipulation of objects is a widely studied task aimed to improve robotic dexterous manipulation capabilities, e.g., \cite{Andrychowicz2020-gc, Qi2023-pr}. Recent works have extended this work further to include tracking the object's pose and reconstructing its geometry during manipulation \cite{Suresh2024-wi}, this was done through the use of RGB-D cameras and vision-based tactile sensors retrofitted into the fingers of the robotic hand.  Building on this foundation, our work aims to further extend in-hand manipulation by introducing real-time semantic labeling, which includes the extraction of semantic properties through the use of the vision-based tactile sensors and fusing this information with the existing geometric map. An overview of our approach can be seen in Figure \ref{fig:overview}.

There are many types of sensors that can provide useful perceptual data for a robot during in-hand manipulation tasks. For instance, RGB-D cameras, which capture both color (RGB) and depth information, have become a popular choice for 3D perception. They enable robots to perceive the environment in three dimensions, and they provide a global sense of an object's appearance and spatial location. However, visual perception has its limitations, particularly in scenarios where the object is partially or fully occluded. This can happen when portions of the object is hidden from the camera view, or under bad lighting situations.

Tactile sensors, on the other hand, can provide direct or indirect information about texture, stiffness, and the forces exerted during physical contact. High-resolution vision-based tactile sensors like the Digit sensor \cite{Lambeta2020-ha} have been helpful in improving robotic perceptual capabilities.

Although the task of pose tracking and shape estimation have been extensively studied in the literature, the ability to extend this task with semantic labeling of objects, especially material classification with tactile perception during in-hand manipulation remains under explored. For example, NeuralFeels \cite{Suresh2024-wi} combines vision and tactile sensing to estimate pose and shape of an object during in-hand manipulation, but does not incorporate semantic labeling into its framework. Moreover, tactile sensing, which holds significant promise for material classification, has received considerably less attention compared to vision-based approaches. This work aims to address these gaps by focusing on tactile-based material classification during in-hand robotic manipulation and integrating these semantic labels into a framework like NeuralFeels.


\section{Related Works}
\label{sec:related}
\textbf{Material Classification with Vision-Based Tactile Sensors}
Vision-based tactile sensors have been used in various material and texture classification applications. In fabrics classification, \cite{Yuan2018-im} fine-tuned a VGG19 model to recognize 11 different material properties of clothing (e.g. thickness, fuzziness, softness, etc.) from tactile data obtained from GelSight tactile images. The results showed that they can predict the properties with a precision much better than chance. They also experimented with other CNN architectures for
the multi-label classification including AlexNet and VGG16, but found that the results were unsatisfactory. \cite{Cao2020-xs} proposed a Spatio-Temporal Attention Model (STAM) that learns where and when to focus on tactile inputs. They showed an improved texture classification accuracy over CNN based non-attention approach. Recently, \cite{Bohm2024-ur} showed reaching 95.2\% classification accuracy of fabrics after 20 epochs of training Inception-V3 model. Other applications involve fossil classification and hardness classification. \cite{Li2024-cx} provides a comprehensive review of vision-based sensors and their applications.

\textbf{Semantic Labeling in Robotic Perception}
Traditional approaches to semantic labeling in robotics often used hand-crafted features or early machine learning techniques, which were effective in structured and static environments but struggled in dynamic or open-world scenarios. Early frameworks, such as those based on feature descriptors like Scale-Invariant Feature Transform (SIFT) \cite{Cortes1995-pq} or Histogram of Oriented Gradients (HOG) \cite{Dalal2005-mb}, combined with classifiers such as Support Vector Machines (SVMs) \cite{Cortes1995-pq}, were foundational in object recognition tasks.

With the advent of CNNs, methods like Fully Convolutional Networks (FCNs) \cite{Long2014-hq} and frameworks such as Mask R-CNN \cite{He2017-jj} significantly advanced semantic segmentation in robotics. Techniques such as SemanticFusion \cite{McCormac2016-iq} augmented SLAM systems by combining 3D reconstruction with semantic labeling. Despite these advances, existing methods often struggled with sparse or noisy annotations and required substantial computational resources for training and inference.

Implicit neural representations have emerged as a powerful tool for scene understanding, allowing the joint encoding of geometry, appearance, and semantics in a unified framework. Neural Radiance Fields (NeRF) \cite{Mildenhall2020-br} demonstrated the capability of neural fields to reconstruct photorealistic 3D scenes from 2D images by learning a continuous function of scene geometry and radiance. \cite{Zhi2021-mw} extend NeRF to include semantic labeling, enabling the joint representation of semantics, appearance, and geometry. Their method allows sparse, in-place annotations to propagate across the scene using the intrinsic multi-view consistency and smoothness properties of NeRF. On the other hand, \cite{Zhi2021-kw} present iLabel, a real-time interactive labeling system that trains a multilayer perceptron (MLP) from scratch during scene reconstruction. This system enables users to define semantic classes on the fly, operating in an open-set manner without requiring pre-training on large datasets.

Our work differs from the reviewed related works in several significant ways. While prior research, such as \cite{Zhi2021-mw} and \cite{Zhi2021-kw}, focuses on incorporating semantic labeling into implicit neural representations for scene understanding, these approaches primarily rely on visual inputs and are tailored for room-scale or large-scale scene reconstruction. In contrast, our method addresses the unique challenges of visuo-tactile perception in robotic manipulation, where inputs are sparse, localized, and dynamic due to in-hand object interactions. Moreover, unlike existing systems, which rely solely on visual data, our method uses tactile sensing into the material classification task.


\section{Our Approach}
\label{sec:approach}
\subsection{Hardware Setup}

\begin{figure}
    \centering
    \includegraphics[width=1.0\linewidth]{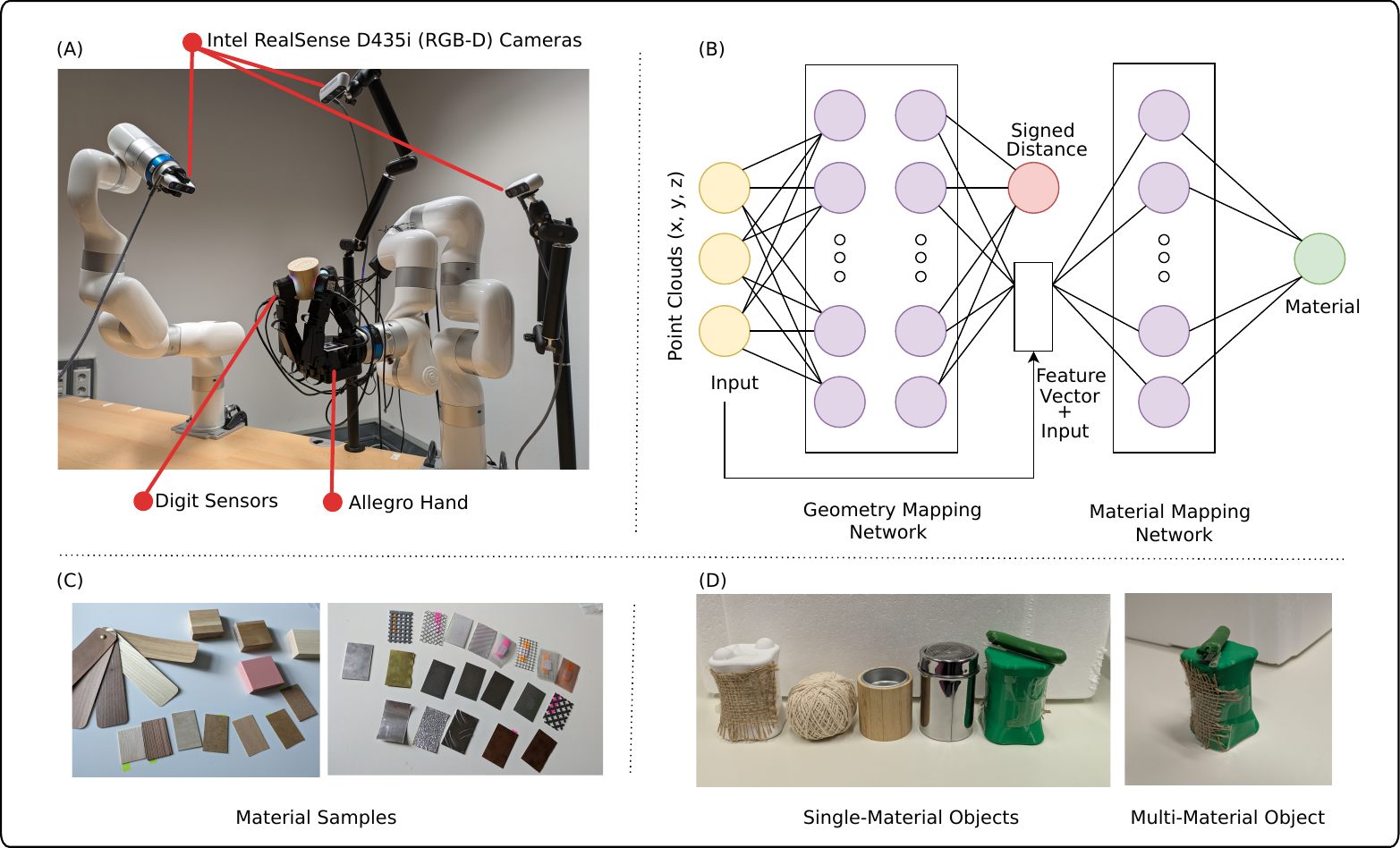}
    \caption{Hardware and backend setup.
(A) Allegro Hand with fingertip Digit sensors Intel RealSense D435i cameras \cite{Unknown2019-sg}.
(B) Dual-branch network: Extended NeuralFeels with a material mapping network. (C) Material Samples used for training the material classification model. (D) Objects used for real-time experiments. From left to right: A toy completely covered with fabric, a yarn ball, wooden candle holder, a stainless steel powder shaker, a plastic toy, and a plastic toy covered partially with a piece of fabric.}
    \label{fig:setup}
\end{figure}

The hardware system is centered around a multi-fingered robotic hand equipped with tactile sensors and multiple RGB-D cameras positioned to provide good visual coverage, as shown in Figure \ref{fig:setup}A. 

In the original NeuralFeels paper, a set of three Intel RealSense D435i cameras were used to capture RGB-D data. These cameras are arranged to provide different viewing angles of the object being manipulated. The cameras can be positioned as follows:
\begin{itemize}
    \item \textbf{Front-Left:} Captures the object and hand from an angle that combines front and side perspectives.
    \item \textbf{Back-Right:} Covers the rear and opposite side of the object and hand.
    \item \textbf{Top-Down:} Offers a bird’s-eye view of the manipulation workspace.
\end{itemize}
In our work, the front-left camera was enough. Intrinsic calibration is performed using the Kalibr tool \cite{Paul-Furgale-Hannes-Sommer-Jerome-Maye-Jorn-Rehder-Thomas-Schneider-email-Luc-OthUnknown-wh}. Intrinsic calibration determines the internal parameters of each camera, such as focal lengths, optical centers, and lens distortions.

The core of the manipulation system is the Allegro Hand \cite{RoboticsUnknown-gk}, a low-cost and dexterous robotic hand with four fingers and 16 independent current-controlled joints, each capable of independent motion. This hand is fitted with four Digit sensors, one on each fingertip, which provide the hand with tactile sensing during interactions with objects. The digit sensors provide high-resolution feedback about the surfaces they touch, capturing information such as texture and compliance.

To control the Allegro Hand, we use a reinforcement learning-based adaptive controller HORA (In-Hand Object Rotation via Rapid Motor Adaptation) \cite{Qi2023-pr}. HORA is designed to facilitate smooth and reliable in-hand object rotation by adapting to the object's physical properties using only proprioceptive feedback. The controller was trained entirely in simulation using cylindrical objects, yet it transfers very well to the real robotic hand and can handle a wide variety of objects with different shapes, sizes, and weights.

\subsection{Framework Overview}

Our framework builds on the NeuralFeels framework and introduces two key components: material classification and material-aware neural implicit representation. These components are integrated to enable semantic labeling of objects with material information during in-hand manipulation tasks.
The material classification task focuses on semantic feature extraction from tactile inputs. 

For material classification we use tactile output images and pass them to a pretrained EfficientNet-B0 \cite{Tan2019-ag} model, which outputs material classification logits. The second task involves building a semantic map by extending the existing neural SDF network, which is used to map the geometric information. 
\fig{fig:overview} shows the general workflow of our approach.

\subsection{Dataset and Data Distribution}
\label{sec:dataset}

A total of 20,749 tactile images were collected from four Digit sensors (index, middle, ring, and thumb), with balanced representation across four material types: plastic, metal, fabric, and wood. Each sensor contributed 1,376 plastic, 1,297 metal, 1,297 fabric, and approximately 1,217 wood samples. The dataset was split into 70\% training, 15\% validation, and 15\% test sets. Samples of the data set used to train the tactile material classifier are shown in \fig{fig:setup}C.

\subsection{Material Classification Model}
\label{sec:material_classification}

The model architecture chosen was EfficientNet-B0, trained using Digit tactile images resized to 224×224 pixels, a batch size of 32, a learning rate of 0.0003, and a weight decay of \( 10^{-5} \). Early stopping was applied with a patience of 10 epochs to prevent overfitting. Data augmentation techniques, such as random rotations and lighting variations, are applied to improve the model’s generalization. During deployment, material classification model processes the images captured by the sensors in real-time. This allows the robot to infer material properties dynamically as it interacts with the object.

\subsection{Semantic Mapping}
After material classification, the second step in our implementation integrates material information into a neural implicit representation. This is achieved by extending the Neural Signed Distance Field (SDF) model to encode material properties alongside geometric information. This extension enables the model to output semantic labels as part of its implicit representation.

 In NeuralFeels, the original network maps 3D inputs (point cloud positions) to a 1-dimensional output (signed distances) using a fully fused MLP with 2 hidden layers of size 64, and ReLU activations. Input encoding is handled using a HashGrid scheme with 16 levels, 2 features per level, a base resolution of 16, a per-level scale of 1.3819, and a log2 hash map size of 24. We extend this network by adding a secondary branch for material mapping, which operates in parallel and receives the same 3D input, with the option of also concatenating a feature vector extracted from the last layer of the SDF network. It uses a spherical harmonics encoding (degree 3) and a separate MLP with 1 hidden layer of size 64, producing a 4-dimensional output corresponding to material class scores. 
 The modified model can be represented as:
\begin{equation}
f_\theta(\mathbf{x}) = \left( \text{SDF}(\mathbf{x}),\; g_\phi\left( \left[\mathbf{x} \,\|\, \mathbf{z}(\mathbf{x})\right] \right) \right)
\end{equation}
where:
\begin{itemize}
  \item \( \mathbf{x} \in \mathbb{R}^3 \) is the 3D query point,
  \item \( \text{SDF}(\mathbf{x}) \in \mathbb{R} \) is the signed distance function output,
  \item \( \mathbf{z}(\mathbf{x}) \in \mathbb{R}^d \) is the feature vector from the final hidden layer of the SDF network,
  \item \( \left[\mathbf{x} \,\|\, \mathbf{z}(\mathbf{x})\right] \in \mathbb{R}^{3 + d} \) denotes concatenation of \( \mathbf{x} \) and \( \mathbf{z}(\mathbf{x}) \),
  \item \( g_\phi \) is the material branch MLP that maps the concatenated vector to material class logits in \( \mathbb{R}^4 \).
\end{itemize}
 
The material mapping network is trained using cross-entropy loss, where the target classes are derived from the material classification task.
Figure \ref{fig:setup}B showcases the structure of our approach.


\section{Results and Discussion}
\label{sec:result}
We conducted multiple real-time experiments using the robot to assess the effectiveness of our approach to semantic labeling during in-hand manipulation. The evaluations focused on how well the system can classify and map material types based on tactile feedback under dynamic conditions. Three key studies were performed: First, an analysis of the tactile classification model's training and performance using an offline hand-collected dataset. Second, a real-time study using objects made of a single material. Last, a more complex scenario involving an object composed of multiple materials. 

\subsection{Training the Model}

\begin{figure}
    \centering
    \includegraphics[width=1\linewidth]{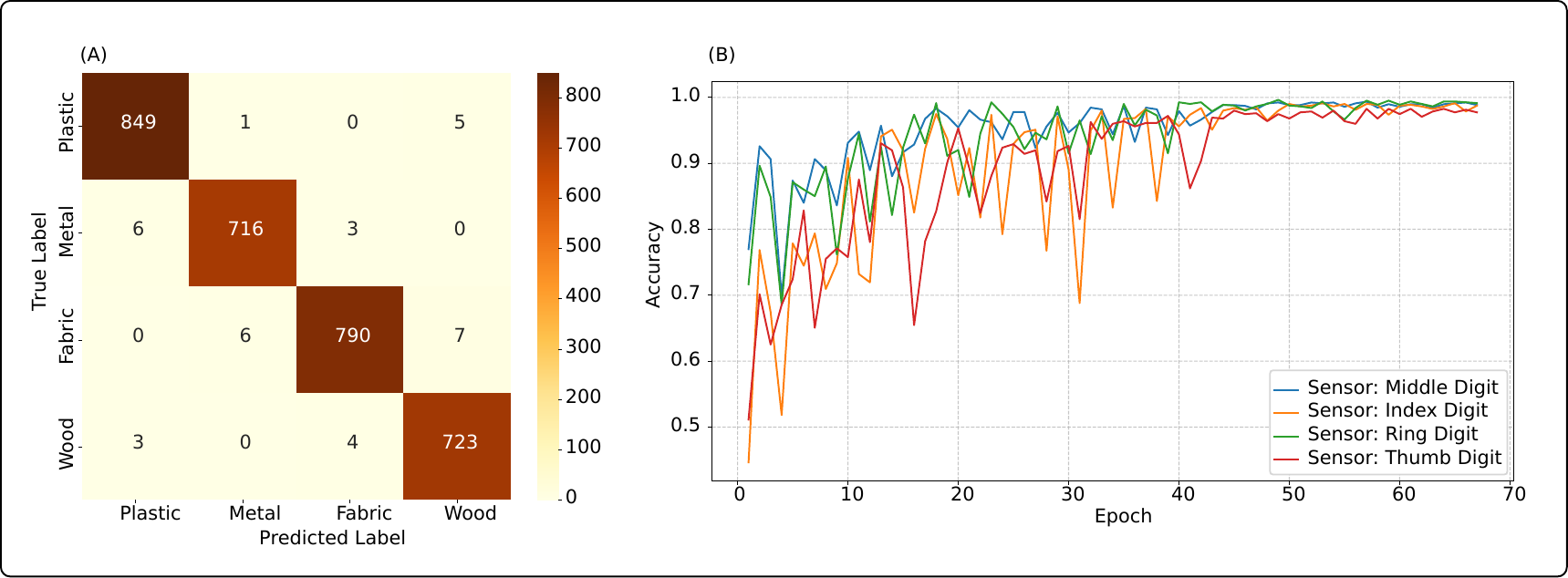}
    \caption{Model performance on hand-collected dataset.
(A) Confusion matrix showing high classification accuracy across four material classes using the hand-collected dataset.
(B) Validation accuracy per sensor, with the ring sensor achieving the highest test accuracy (99.60\%) and the thumb sensor showing relatively lower performance (97.42\%).
These results are based on the offline dataset used for model training and validation, prior to deployment in real-time scenarios.}
    \label{fig:evaluation1}
\end{figure}

The tactile classification model, based on EfficientNet-B0, was trained using a hand-collected dataset of tactile images resized to 224×224 pixels. Training of was carried out with early stopping (patience of 10 epochs), and the best validation accuracy of 99.04\% was achieved at epoch 57. The final training and validation accuracies were 99.15\% and 98.62\%, respectively, with relatively low final losses. On the held-out test set, the model achieved an overall accuracy of 98.88\%. When evaluated per sensor, performance remained consistently high, with the ring sensor achieving the highest accuracy at 99.60\% and the thumb sensor the lowest at 97.42\%. These results confirm the high confidence of the model in distinguishing material classes under controlled offline conditions. 
Our trained model shows high accuracy across all sensors as shown in Figure \ref{fig:evaluation1}.

\subsection{Evaluation on Single Material Objects}

\begin{figure}
    \centering
    \includegraphics[width=1\linewidth]{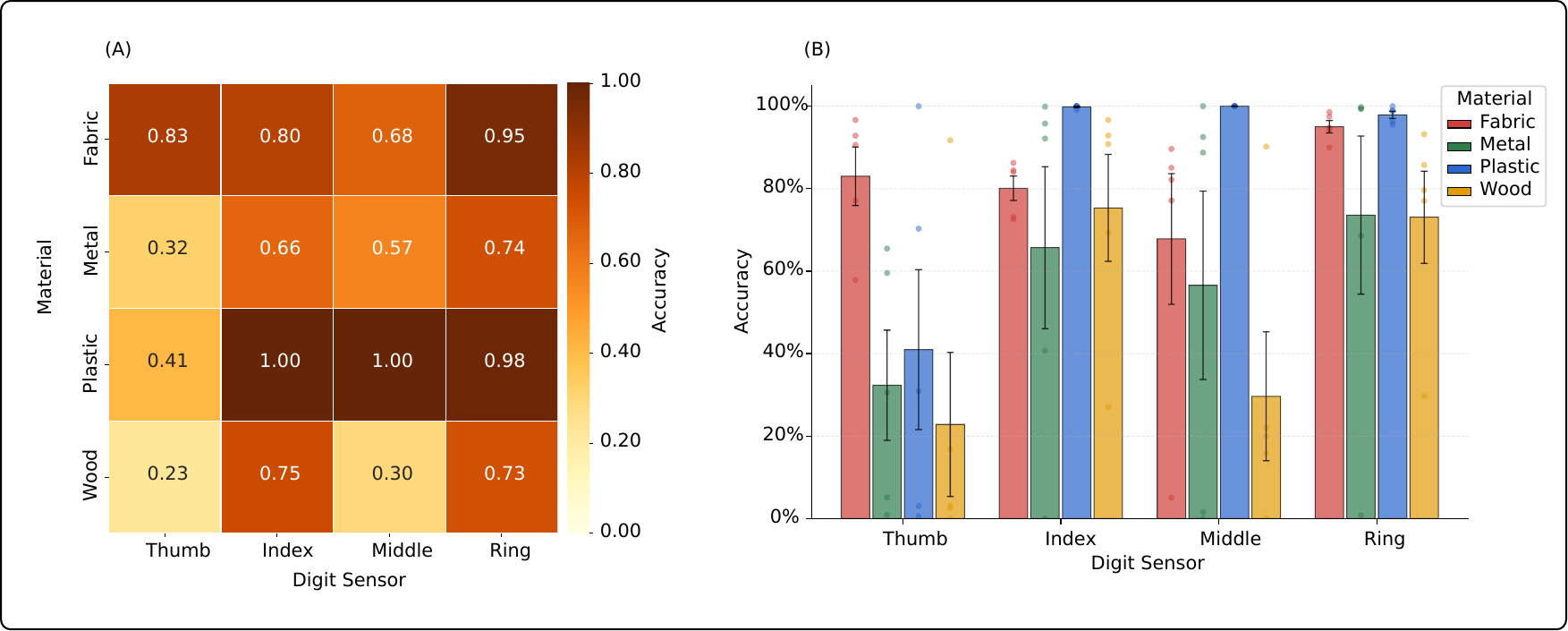}
    \caption{Evaluation of material classification using real-time robot-collected data.
(A) Heatmap showing mean classification accuracy across digit sensors. Plastic materials achieve near-perfect accuracy on all sensors except the thumb. Ring and index fingers yield the highest overall accuracy. Wood and metal are harder to classify, especially with the thumb.
(B) Bar plot of classification accuracy by digit sensor with error bars. Plastic shows high accuracy and low variability (except thumb). Ring and index fingers consistently perform better, while wood and metal exhibit lower and more variable accuracy, particularly on thumb and middle fingers.}
    \label{fig:Single}
\end{figure}

In the single-material object evaluation, the trained model was tested in real-time on objects composed entirely of one material: plastic, metal, fabric, or wood, as shown in Figure \ref{fig:setup}D. Across five trials per object, classification accuracy was measured for each digit sensor. In Figure \ref{fig:Single}, the results are presented.
Plastic materials achieved the highest overall performance, with accuracies approaching 100\% across most sensors. The ring sensor consistently delivered the best results, including 99\%+ accuracy for both plastic and fabric. The reason could be due to the manipulation policy where the ring and the index are always in firm contact with the object to ensure that the object doesn't slip. 

Fabric also showed strong classification rates, particularly on the ring (95\%) and thumb (83\%) sensors. Metal and wood were more challenging; for metal, accuracy ranged from 32\% (thumb) to 74\% (ring), while wood had a low of 23\% (thumb) and peaked at 75\% (index). The thumb and middle sensors generally exhibited the lowest performance, especially for wood and metal. Standard deviation was lowest for plastic and highest for wood.

 The low classification accuracy observed from the thumb sensor, particularly for plastic, wood, and metal materials, can be attributed to the robot's in-hand rotation policy during the runs. This policy often causes the thumb to make less firm or inconsistent contact with the object's surface, resulting in suboptimal tactile images. Consequently, the tactile data captured by the thumb lacks texture information necessary for reliable material classification. In contrast, the ring and index fingers maintain more stable and direct contact during manipulation, leading to higher-quality tactile outputs and more accurate classification.

\subsection{Evaluation on a Multi Material Object}

\begin{figure}
    \centering
    \includegraphics[width=1\linewidth]{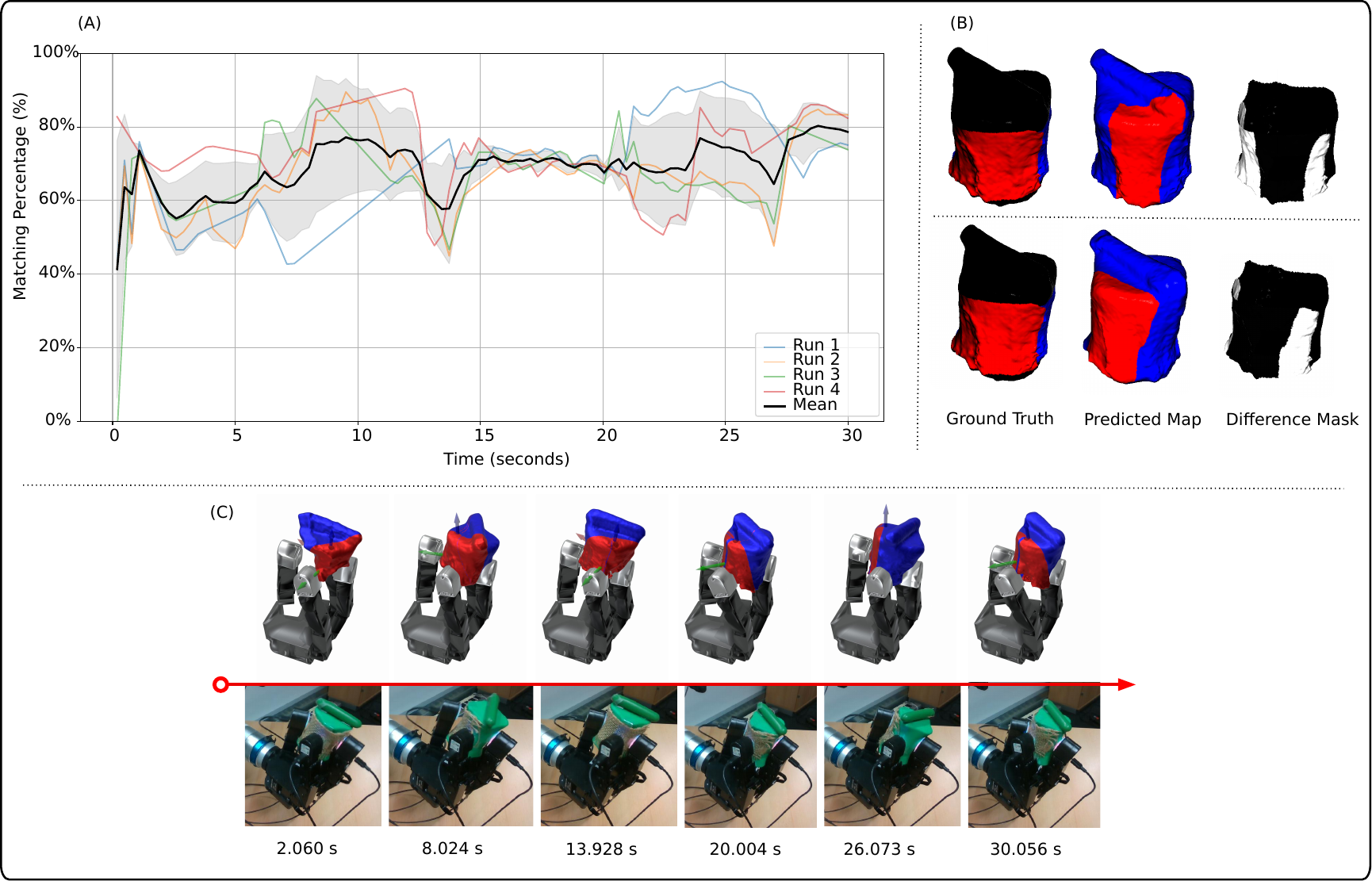}
    \caption{Evaluation on Real-Time Robot-Collected Runs of a Multi-Material Object. (A) Matching percentage across four runs showing progressive improvements in map alignment over time, with final values exceeding 75\% in each case. (B) Two examples of the resulting maps. (Left) The ground truth material map, colored according to material classes with a black color on regions not under study. (Middle) The predicted material map. (Right) The difference mask which is used to calculate the matching percentage. (C) An example run showcasing the progression of the predicted map over time.}
    \label{fig:multi}
\end{figure}

For this study, we used a custom made object consisting of plastic regions and fabric regions, as shown in Figure \ref{fig:setup}D. We applied our framework while the object was manipulated by the in-hand rotation policy for 4 different runs. For each run, a material map gets generated every few time steps, as in Figure \ref{fig:multi}C. The predicted material map is then compared to the ground truth map on the regions of interest as in Figure \ref{fig:multi}B. Afterwards, a matching percentage between the maps can be obtained. Figure \ref{fig:multi}A showcases the matching percentage of the predicted material map to the ground truth.

The matching percentages for each run showed consistent improvements across processing stages. In Run 1, the percentage increased from 43.70\% to 75.09\%. Run 2 began at 43.93\% and rose to 83.32\%. Run 3 started at 47.85\% and reached 75.93\%. Run 4 started with a relatively high initial value of 74.26\% and improved further to 85.12\%. The combined average matching percentage across all runs was 79.87\% with a standard deviation of 4.41\%.

Despite the variability in the results across time, the observed improvements are encouraging. Several factors contribute to the variability, including imperfect classifier accuracy, continuous object motion during manipulation, and the limitations of tactile sensing, particularly in capturing fine texture details due to inconsistent contact. Nevertheless, achieving an average matching accuracy of 79.87\% across runs demonstrates the effectiveness of the approach.


\section{Conclusion}
\label{sec:conclusion}
This work presents a method for semantic labeling of objects during in-hand robotic manipulation using tactile sensing. We use implicit neural representations to encode both geometric and material properties. 

We demonstrate the feasibility of our approach. By training a tactile image-based classification model on an offline dataset and deploying it in real-time scenarios, we evaluated its performance on both single-material and multi-material objects. The system achieved high accuracy across multiple digit sensors, particularly the ring and index fingers, and showed reliable performance in distinguishing between materials such as plastic, fabric, wood, and metal. The multi-material experiments further validated the framework's ability to construct material maps over time, achieving an average matching accuracy of 79.87\% across several manipulation runs.

While the results are promising, the study also highlights several practical limitations related to contact consistency, sensor coverage, and material diversity. Variability in performance, especially for more challenging materials and under dynamic conditions, suggests the need for improved contact strategies and broader training data. Future work could explore adaptive manipulation policies, expanded material sets, and integration with other sensing modalities to further enhance robustness and generalization. Overall, this work provides a strong foundation for tactile-based semantic understanding of objects in robotic manipulation tasks.


\clearpage
\section{Limitations and Future Work}

While the proposed framework demonstrates strong performance in real-time material classification and mapping, several limitations remain:

\begin{itemize}
    \item The in-hand rotation policy does not ensure stable or firm contact across all fingers, particularly the thumb. This leads to variability in tactile image quality and reduces classification reliability for certain materials.
    
    \item The study included only four material types (plastic, fabric, wood, metal). Broader generalization requires expanding to a more diverse set of materials with varied textures and stiffness levels.
    
    \item Tactile data is highly sensitive to contact angle, pressure distribution, and motion, all of which can vary unpredictably during dynamic manipulation.
    
    \item The rotation policy is predefined and does not adapt based on object shape or sensor feedback. This limits optimal sensor coverage across different objects.
    
    \item This work is limited to tactile input. Incorporating other modalities, such as vision or force sensing, could enhance robustness and resolve cases where tactile data alone is insufficient.
\end{itemize}

Future work could focus on developing adaptive manipulation strategies that dynamically adjust contact points based on feedback from tactile sensors. Expanding the material dataset to include a broader range of textures and mechanical properties will help improve model generalization. Additionally, integrating vision and proprioceptive sensing could provide complementary information, especially for multi-material and irregularly shaped objects. Online learning techniques and domain adaptation methods could also be explored to close the performance gap between offline training and real-time deployment. Finally, efforts should be made to apply the system to more complex manipulation tasks, such as object sorting or tool use, to demonstrate its utility in practical applications.



\bibliography{paperpile}  

\end{document}